\documentclass{article}

\usepackage{arxiv}
\usepackage{microtype}
\usepackage{graphicx}
\usepackage{multirow}
\usepackage{booktabs} 
\usepackage{subcaption}
\usepackage{algorithm}
\usepackage{algorithmic}
\usepackage[numbers]{natbib}

\usepackage{hyperref}

\usepackage{amsmath,amsfonts,amsthm,mathabx,amssymb}

\usepackage[english]{babel}
\usepackage{amsthm}

\theoremstyle{case}

\usepackage{xcolor}
\usepackage{tikz}
\usepackage{subcaption}
\usepackage{adjustbox}
\usetikzlibrary{positioning}
\tikzset{
	layer/.style={draw, minimum width=4cm, minimum height=1cm, align=center},
	resblock/.style={draw, dashed, minimum width=4cm, minimum height=1cm, align=center},
	arrow/.style={->, thick}
}

\usepackage{siunitx}

\usepackage{xspace}
\newcommand{\PkgName}{DeepXube}

\usepackage{multirow}
\usepackage{subcaption}
\usepackage{booktabs}       

\usepackage{longtable}

\title{The \PkgName{} Software Package for Solving Pathfinding Problems with Learned Heuristic Functions and Search}

\author{
	Forest Agostinelli\\
	Department of Computer Science and Engineering\\
	University of South Carolina\\
	\texttt{foresta@cse.sc.edu} \\
}
\date{}

\begin{document}

\maketitle

\begin{abstract}
\PkgName{} is a free and open-source Python package and command-line tool that seeks to automate the solution of pathfinding problems by using machine learning to learn heuristic functions that guide heuristic search algorithms tailored to deep neural networks (DNNs). \PkgName{} is comprised of the latest advances in deep reinforcement learning, heuristic search, and formal logic for solving pathfinding problems. This includes limited-horizon Bellman-based learning, hindsight experience replay, batched heuristic search, and specifying goals with answer-set programming. A robust multiple-inheritance structure simplifies the definition of pathfinding domains and the generation of training data. Training heuristic functions is made efficient through the automatic parallelization of the generation of training data across central processing units (CPUs) and reinforcement learning updates across graphics processing units (GPUs). Pathfinding algorithms that take advantage of the parallelism of GPUs and DNN architectures, such as batch weighted A* and Q* search and beam search are easily employed to solve pathfinding problems through command-line arguments. Finally, several convenient features for visualization, code profiling, and progress monitoring during training and solving are available. The GitHub repository is publicly available at \url{https://github.com/forestagostinelli/deepxube}.
\end{abstract}

\section{Introduction}
Pathfinding problems are found throughout mathematics, chemistry, robotics, and computing. Pathfinding algorithms that solve such problems include those based on human-designed algorithms, pattern databases \cite{korf_1997,culberson1998pattern}, and heuristics derived from formal representations of pathfinding domains \cite{helmert2006fast}. In the past decade, it has been shown that representing heuristic functions with deep neural networks (DNNs) \cite{schmidhuber2015deep,lecun2015deep}, training them with machine learning, and combining them with heuristic search can rival or outperform the aforementioned methods while assuming a black-box representation of the pathfinding domain and without relying on pre-existing solvers \cite{agostinelli2019solving,chen2020retro,muppasani2024comparing,agostinelli2024specifying,chervovmachine}. These techniques have been applied to puzzle solving \cite{agostinelli2019solving,chervovmachine}, chemical reaction mechanism pathfinding \cite{panta2024finding}, quantum algorithm compiling \cite{zhang2020topological,bao2024twisty,qiuhao2024efficient,turner2025quantum}, cryptography \cite{jin20203d}, and parking lot optimization \cite{siddique2021puzzle}.

\begin{figure}[h]
	\centering
	\includegraphics[width=0.45\textwidth]{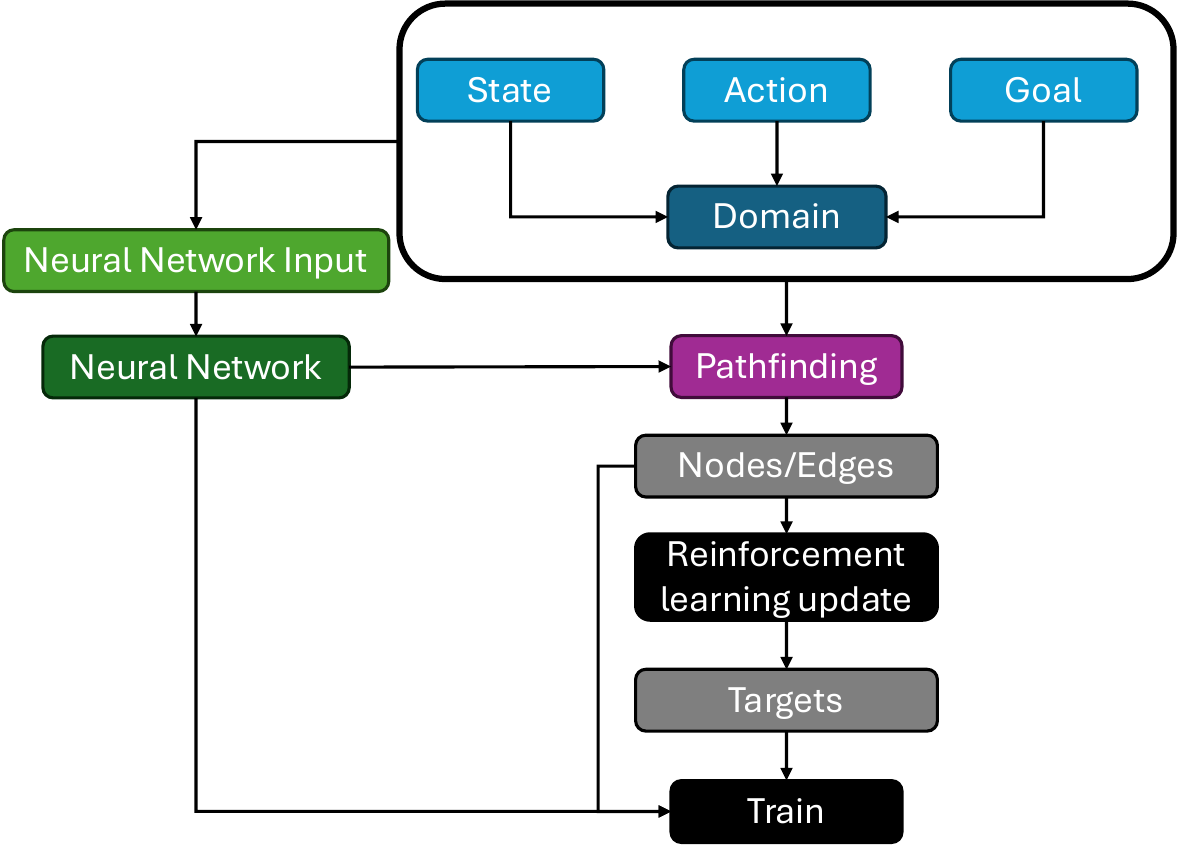}
	\caption{An overview of how \PkgName{} trains heuristic functions. The user implements the domain, along with its states, actions, and goals, as well as a DNN architecture that processes states, goals, and possibly actions. \PkgName{} then samples problem instances from the domain and uses the DNN to attempt to solve them with pathfinding, generating nodes and edges from the search tree. A reinforcement learning update is then applied to either the nodes or edges to compute targets. The DNN is then trained to match the targets.}
	\label{fig:deepxube}
\end{figure}

\PkgName{} (pronounced ``Deep Cube'') is a free and open-source Python package that seeks to automate the solution of pathfinding problems given only a black-box implementation of a pathfinding domain and a heuristic function represented as a DNN. \PkgName{} is comprised of the latest advances in the combination of deep reinforcement learning, supervised learning, and heuristic search for pathfinding. These include batch weighted A* and Q* search \cite{agostinelli2024q}, limited-horizon Bellman-based learning (LHBL) \cite{hadar2026beyond}, hindsight experience replay \cite{andrychowicz2017hindsight}, supervised learning based on random walk path costs \cite{chervovmachine}, and specifying goals with answer set programming \cite{agostinelli2024specifying, agostinelli2025conflict}. Pathfinding domains are implemented with arbitrary Python code and heuristic functions are implemented with arbitrary PyTorch code \cite{paszke2019pytorch}. Given an implementation of a pathfinding domain and heuristic function, \PkgName{} samples problem instances from the domain and uses pathfinding to generate training data for the heuristic function, as shown in Figure \ref{fig:deepxube}. Training is automatically parallelized across a user-specified number of central processing units (CPUs) and the DNN is parallelized across all visible graphics processing units (GPUs). Several methods to monitor progress are provided, including TensorBoard \cite{abadi2015tensorflow}, a pathfinding-specific training summary, periodic testing of solution quality on a test set, and a log file. After training, given pathfinding problems, \PkgName{} uses the trained heuristic function with heuristic search to solve problems and produces a file summarizing the solutions.

\section{Preliminaries}
\subsection{Pathfinding}
A \textbf{pathfinding domain} is defined as a weighted directed graph \cite{pohl1970heuristic}, where nodes represent states, edges represent actions that transition between states, and weights on the edges represent transition costs. The set of all possible outgoing edges of a state, $s$, which corresponds to the actions that can be taken in $s$, is denoted $E_{o}(s)$. The transition function is represented by $T$, where $s' = T(s,a)$ for some $a \in E_{o}(s)$. The transition cost function is represented by $c$, where $c(s,a)$ is the transition cost when taking action $a$ in state $s$. The set of all possible actions is denoted $\mathcal{A}$. A \textbf{pathfinding problem instance} is defined by a tuple $(D, s_0, g)$, where $D$ is the pathfinding domain, $s_0$ is the start state, and $g$ is the set of states considered goal states. Given a pathfinding problem, the objective is to find a path, which is a sequence of actions, that transforms the start state into a goal state while attempting to minimize the path cost, where the path cost is the sum of transition costs. The cost-to-go is the cost of a shortest path from a state, $s$, to a closest goal state in a goal, $g$, which we will denote $h^*(s,g)$.

\subsection{Heuristic Search}
Heuristic search is a widely used approach for solving pathfinding problems and is guided by a heuristic function that estimates the cost-to-go. Heuristic functions which operate on the nodes of the pathfinding domain graph map a state, $s$, and a goal, $g$, to an estimate $h(s,g) \approx h^*(s,g)$. Heuristic functions which operate on the edges of the pathfinding domain graph map a state, $s$, goal, $g$, and action, $a$, to an estimate $q(s,g,a) \approx c(s,a) + h^*(T(s,a),g)$. We will refer to heuristic functions which operate on nodes as heuristic-v functions, since they are often trained with value iteration \cite{bellman1957dynamic}, and heuristic functions which operate on edges as heuristic-q functions, since they are often trained with q-learning \cite{watkins1992q}. Heuristic search algorithms create a search tree, where nodes represent states and edges represent actions.

\subsubsection{Graph Search}
Graph search algorithms in the context of heuristic search maintain a closed dictionary that maps states to their cheapest path cost observed so far, in that search. The most notable graph search algorithm that uses a heuristic-v function, A* search \cite{hart1968formal}, iteratively selects nodes from the search tree for expansion from a priority queue prioritized by the cost of the node, which is the sum of its path cost from the start node (i.e. cost-to-come computed by the sum of transition costs) and its heuristic value (i.e. cost-to-go computed by $h$). A node is expanded by applying every possible action to the state associated with that node and creating child nodes from the resulting states. The cost of each child node is computed and the child node is added to the priority queue if the state associated with the node is not in the closed dictionary or has been reached via a cheaper path. A* search terminates when a node associated with a goal state is selected for expansion and returns the path to that node.

Q* search \cite{agostinelli2024q} is a variant of A* search that uses a heuristic-q function. Q* search iteratively selects edges for expansion and, in doing so, only generates nodes when removing an edge from the priority queue. As a result, the number of nodes generated at each iteration is independent of the number of available actions. Furthermore, some deep Q-network architectures can map a state, $s$, and a goal, $g$, to $q(s,g,a)$ for all possible actions, $a$ \cite{mnih2015human}, thus computing the heuristic value for all edges with a single forward pass through a DNN.

Both A* and Q* search have batched and weighted variants referred to as batch weighted A* search (BWAS) and batch weighted Q* search (BWQS), respectively. For weighted search \cite{pohl1970heuristic}, the path cost is multiplied by a scalar, $\lambda \in [0,1)$. This decreases the influence the path cost has on the overall cost and exchanges potentially faster search times for potentially greater solution costs. For batched search, $B > 1$ nodes or edges are removed from the priority queue each iteration. When heuristic functions are represented as DNNs, this exploits the parallelism provided by GPUs for DNNs.

\subsubsection{Beam Search}
As opposed to graph search, beam search neither maintains a closed dictionary nor uses a priority queue. Beam search maintains a ``beam'' of nodes of a maximum size, $B$, starting with just the start node in the beam. At each iteration, beam search selects the top $B$ edges for each node in the beam and transitions along those edges to produce the nodes in the beam for the next iteration. The metric to evaluate the edges could come from computing $-(c(s,a) + h(T(s,a),g))$ with a heuristic-v function, computing $-q(s,g,a)$ with a heuristic-q function, or computing the probability density for each action with a policy function, $\pi(a|s,g)$. Stochasticity can be added to the beam search by selecting edges according to a Boltzmann distribution with some given temperature, $\tau$, and/or randomly selecting edges with some probability $\epsilon$. One notable special case is the $\epsilon$-greedy policy, often used in reinforcement learning, where $B=1$ and the action taken is random with probability $\epsilon$ and greedy, otherwise (i.e. $\tau=0$).

\subsection{Learning Heuristic Functions}
Reinforcement learning can be used to learn heuristic functions based only on a definition of the domain and a way to generate problem instances. A DNN used to represent a heuristic-v function, $h_\theta$, with parameters, $\theta$, can be trained using approximate value iteration \cite{bellman1957dynamic,bertsekas1996neuro,sutton2018reinforcement}. The DNN is trained with gradient descent to minimize the loss function in Equation \ref{eq:viloss}, where $N$ denotes the batch size and $\theta$ denotes the parameters of the DNN. The value iteration update in the context of pathfinding is shown in Equation \ref{eq:viup}, where $\theta^-$ are the parameters of the target network \cite{mnih2015human} which are periodically updated to $\theta$.

\begin{equation} \label{eq:viloss} 
	L({\theta}) = \frac{1}{N}\sum_{i}^N{(h'(s_i,g_i) - h_{\theta}(s_i,g_i))^2} 
\end{equation}

\begin{equation} \label{eq:viup} 
	h'(s,g) =
	\begin{cases}
		0, & \text{if } s \in g, \\
		\min\limits_{a \in E_o(s)}{c(s,a) + h_{\theta^-}(T(s,a),g)}, & \text{ow}.
	\end{cases}
\end{equation}

\noindent A DNN used to represent a heuristic-q function, $q_\theta$, with parameters, $\theta$, can be trained using approximate Q-learning \cite{watkins1992q,mnih2015human}. The DNN is trained with gradient descent to minimize the loss function in Equation \ref{eq:qloss}. The q-learning update in the context of pathfinding is shown in Equation \ref{eq:qup}, where $s' = T(s,a)$.

\begin{equation} \label{eq:qloss} 
	L({\theta}) = \frac{1}{N}\sum_{i}^N{(q'(s_i,g_i,a_i) - q_{\theta}(s_i,g_i,a_i))^2} 
\end{equation}

\begin{equation} \label{eq:qup} 
	q'(s,g,a) =
	\begin{cases}
		0, & \text{if } s \in g, \\
		c(s,a) + \min\limits_{a' \in E_o(s')}{q_{\theta^-}(s',g,a')}, & \text{ow}.
	\end{cases}
\end{equation}

\section{Defining Domains}
To define a domain with \PkgName{}, the following methods must be implemented: sampling problem instances (\verb|samp_prob_insts|), sampling a valid action in a given state (\verb|samp_state_act|), the state transition and transition cost when taking a given action in a given state (\verb|next_state|), and a goal test given a state and goal (\verb|is_solved|). The benefit of assuming the domain definition is a black box is that its implementation can depend on arbitrary Python packages, such as RDKit for chemistry \cite{landrum2013rdkit}, and goals can represent a set of states without having to know of any state that is a member of that set. For example, for chemical synthesis, states can be molecules and goals can be properties that a molecule should have, implicitly defining the set of molecules that have those properties. For quantum circuit synthesis, states could be quantum circuits and goals could be matrices representing quantum algorithms, implicitly defining the set of circuits that implement those algorithms \cite{turner2025quantum}.

\subsection{Mixins}
\PkgName{} contains multiple subclasses of the \verb|Domain|, which we refer to as ``mixin'' classes, that define additional functionality. Depending on the pathfinding or training algorithm used, the domain may need to subclass certain mixins. For example, A* search in an enumerable action space requires the domain to subclass the \verb|ActsEnum| mixin from which the result of applying all possible actions to a state can be obtained with the \verb|expand| method. Generating problem instances with random walks requires the domain to subclass the \verb|GoalSampleableFromState| mixin from which goals can be obtained through the \verb|samp_goal_from_state| method. The mixin classes also simplify domain definition by automatically implementing methods based on the implementation of simpler methods. For example, if the action space is enumerable and fixed, the child class can implement a single method that gets all possible actions and the mixin class implements \verb|samp_state_act| as well as \verb|expand|. A domain can inherit from multiple mixins and any automatically implemented method can be overridden, if the user deems it necessary.

\begin{figure*}[h]
	\centering
	\includegraphics[width=0.95\textwidth]{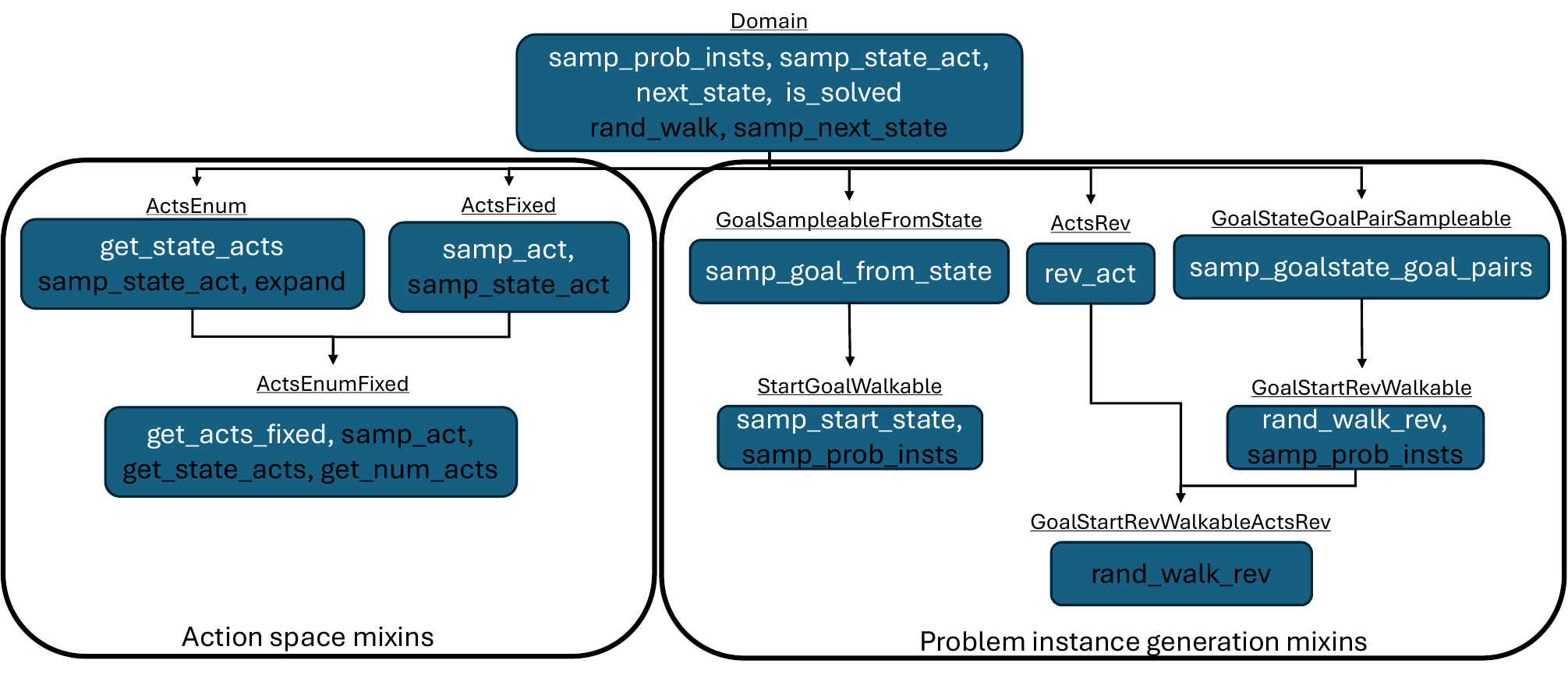}
	\caption{The primary mixin classes used to construct domains. Methods in white are abstract and methods in black are implemented by the class. Mixins inherit all functionality of their ancestors.}
	\label{fig:domain}
\end{figure*}

\begin{figure}[h]
	\centering
	\includegraphics[width=0.45\textwidth]{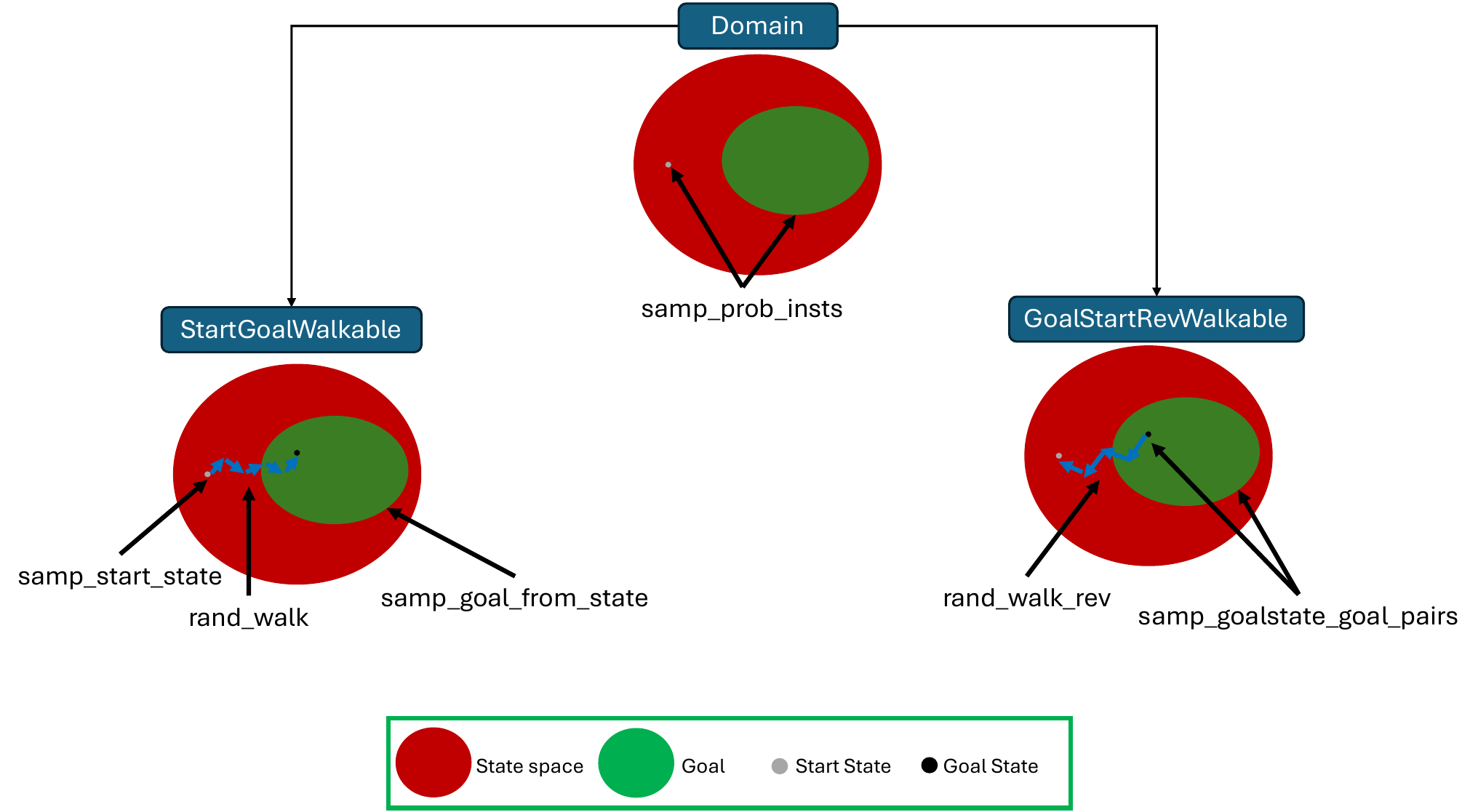}
	\caption{A visualization of the problem instance generation function along with two mixin classes for generating problem instances by either 1) sampling a start state, taking a random walk, and sampling a goal from the terminal state of that random walk; or 2) sampling a goal and a corresponding goal state, taking a random walk in reverse, and using the terminal state of that random walk as the start state.}
	\label{fig:probinstgen}
\end{figure}

While sampling problem instances may be implemented in any manner, \PkgName{} adds mixin classes that define ways of sampling problem instances that can work well in practice. For actions that can be reversed, such as the Rubik's cube, sliding tile puzzles, and certain quantum circuit synthesis tasks, sampling a goal state and corresponding goal and taking a random walk in reverse from the goal state has worked well in practice \cite{agostinelli2019solving, zhang2020topological, chervovmachine}. A reverse walk can also be done for irreversible action spaces given an implementation of a reverse walk, such as Sokoban where boxes are pulled instead of pushed. For action spaces that are not easily reversed, such as many robotics tasks and chemical reaction mechanisms, or domains where sampling a reachable goal is difficult, sampling a start state, taking a random walk, and sampling a goal from the terminal state has worked well in practice \cite{agostinelli2024specifying,panta2024finding,agostinelli2024learning}. The mixin classes are summarized in Figure \ref{fig:domain}. Visualizations of problem instance generation mixins are shown in Figure \ref{fig:probinstgen}.

Two mixins, \verb|StateGoalVizable| and \verb|StringToAct| are provided to visualize problem instances as well as use the terminal to apply actions to the states and visualize the resulting next states. The visualization is done via a \verb|matplotlib| \verb|Figure| object, so visualizations can be interactive, if need be. For example, a visualization for the Rubik's cube could allow the user to click and drag to rotate the cube.

\section{Defining Neural Networks}
\PkgName{} has strong integration with both heuristic-v and heuristic-q functions and preliminary integration with policy functions for continuous action spaces, such as those seen in robotics, or very large action spaces where enumeration is impractical, such as multi-agent path finding. To ensure the implemented DNN accepts the appropriate input, the DNN class specifies what \verb|NNetInput| class it requires. The \verb|NNetInput| class converts the anticipated Python objects to a representation suitable for the DNN.

\begin{figure}[h]
	\centering
	\includegraphics[width=0.45\textwidth]{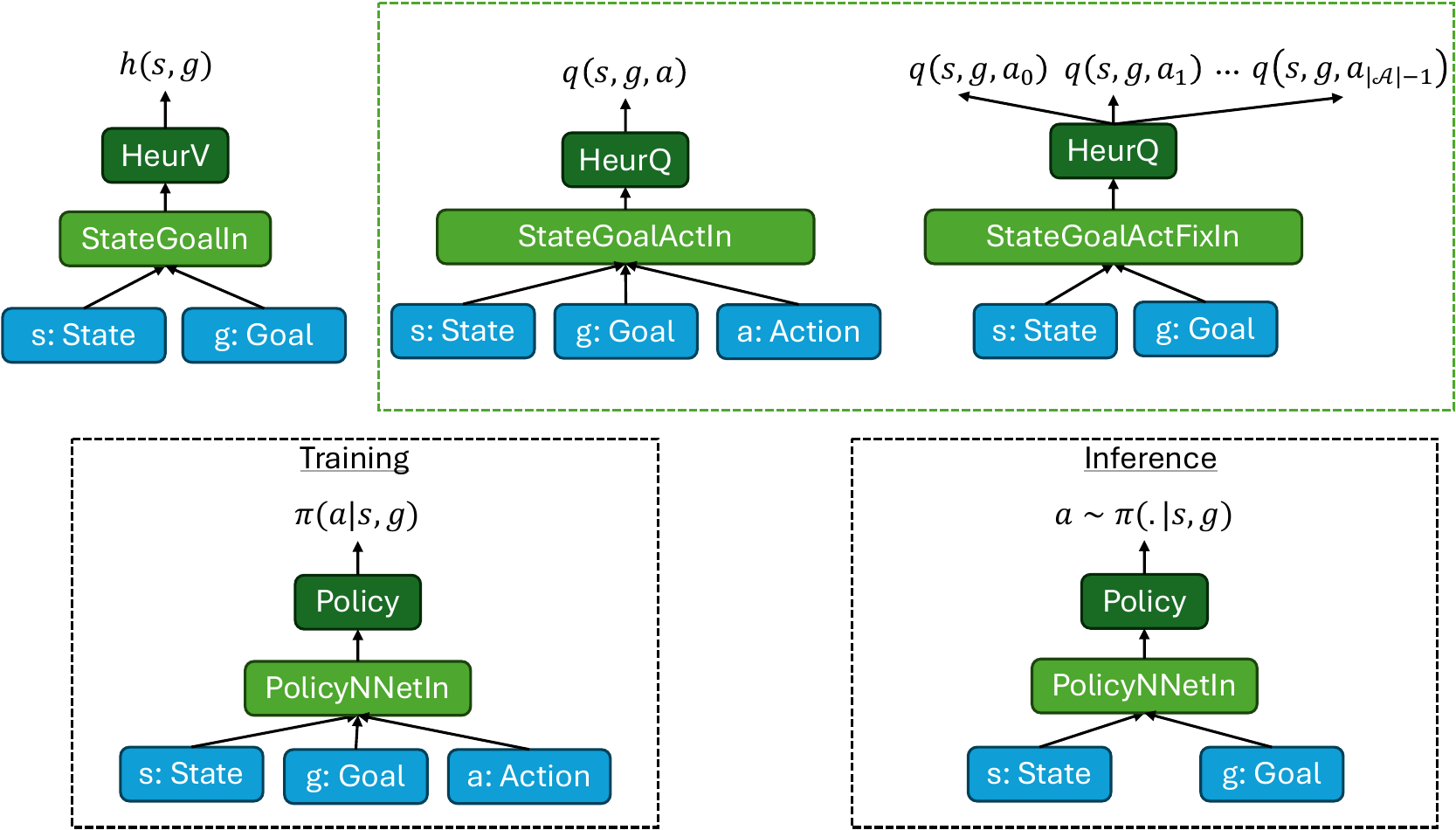}
	\caption{The interaction between states, goals, actions, neural network inputs, and heuristic/policy functions. Heuristic-q functions take an action as an input or output the q-values for all possible actions. A policy function takes an action as an input during training and samples actions during inference.}
	\label{fig:nnets}
\end{figure}

Heuristic-v functions take a state and goal as an input. Heuristic-q functions can either take a state, goal, and action as an input, or take a state and goal as an input and output the heuristic values for all possible edges with a single forward pass. Currently, this is only supported for a fixed action space; however, future updates plan to implement this for domains whose action spaces are dynamic and can be represented with a structured output, such as graphs \cite{you2018graph}. For policy functions, during training, the DNN is trained to match the distribution of edges seen during heuristic search, with the intuition that successful searches will be biased towards edges that lead to the goal \cite{mark2024policy}. During inference, the policy network samples actions from its modeled distribution. However, the approach to learning a policy for pathfinding is preliminary and may be significantly updated in the future. A visualization of the neural networks that can be defined along with their inputs and outputs is shown in Figure \ref{fig:nnets}.

There can be multiple \verb|NNetInput| classes for different kinds of DNN architectures for a single domain and single kind of heuristic function. For example, a domain could have a \verb|NNetInput| for a heuristic-v function represented as a convolutional neural network and a  \verb|NNetInput| for a heuristic-v function represented as a transformer \cite{vaswani2017attention}. This allows architecture search for a single domain over multiple different kinds of architectures as well as multiple domains to use the same DNN architecture provided the appropriate \verb|NNetInput| class is defined.

\section{Pathfinding Algorithms}
\begin{figure*}[h]
	\centering
	\includegraphics[width=\textwidth]{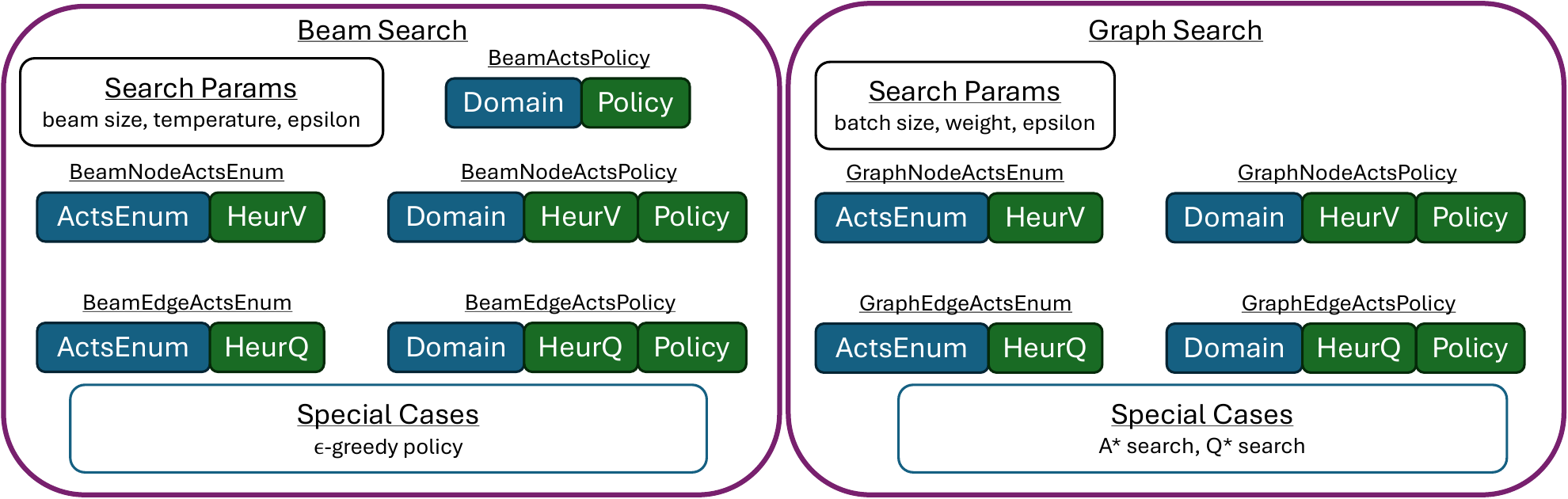}
	\caption{The available pathfinding algorithms. The mixin class that the domain must subclass is in blue and the functions used are in green. If both a heuristic and policy function are used, then the pathfinding algorithm is guided by the heuristic function and the policy function is used to sample actions.}
	\label{fig:pathfind}
\end{figure*}

\PkgName{} has pathfinding algorithms that search over both nodes and edges, that operate on enumerable and infinite action spaces, and that perform beam search or graph search. Pathfinding algorithms specify what mixin class a domain must subclass as well as what functions it uses. For pathfinding algorithms that operate on enumerable action spaces, the domain must subclass \verb|ActsEnum|. Otherwise, the pathfinding algorithms accept any subclass of \verb|Domain| while also requiring a policy function to sample actions. For beam search, a temperature of $0$ means that nodes are selected greedily unless $\epsilon$ is greater than 0. Graph search is also given an $\epsilon$ parameter where a random node or edge is popped from the priority queue with probability $\epsilon$. This is primarily used for exploration during training. A summary of the pathfinding algorithms is shown in Figure \ref{fig:pathfind}. Note that, while the methods for training policy functions may change, the pathfinding algorithms that expect policy functions only assume the ability to sample actions from the policy function and, therefore, are agnostic to how the policy function is trained.

\section{Learning Heuristic Functions}
\begin{figure}[h]
	\centering
	\includegraphics[width=0.45\textwidth]{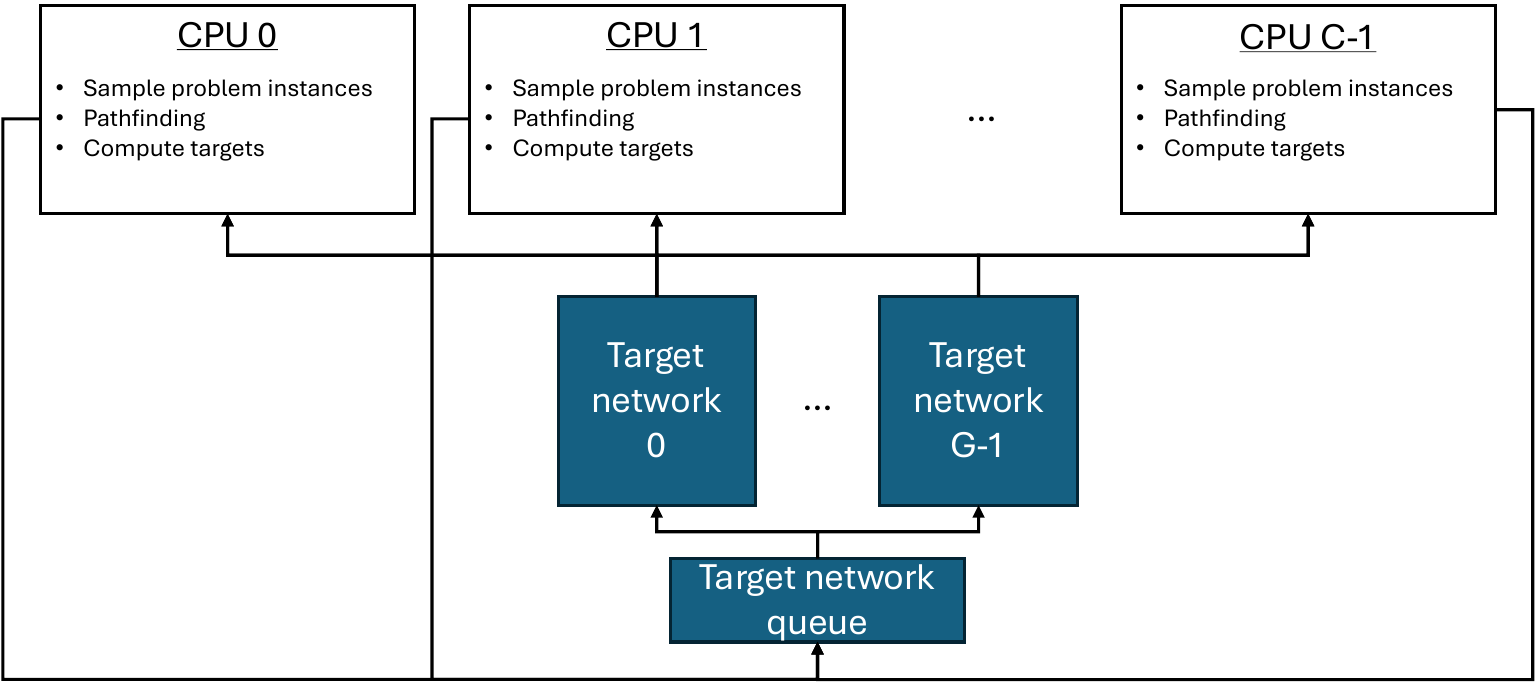}
	\caption{Training is parallelized across $C$ CPUs and $G$ GPUs. CPUs sample problem instances, perform heuristic search, and compute targets for training. When computing targets, the CPU sends the corresponding input data and its process ID to the target network queue, which sends the data to the first available target network. The output of the target network is obtained and sent to the correct CPU via the given process ID.}
	\label{fig:trainmulti}
\end{figure}

When learning heuristic functions, the generation of training data is parallelized across $C$ CPUs and $G$ GPUs and the heuristic function training is parallelized across $G$ GPUs. CPUs sample problem instances, perform pathfinding, and compute the target for the DNN using either value iteration on the nodes or q-learning on the edges produced by pathfinding. The target network can also be used to guide the pathfinding algorithm; however, while this may work for some domains, this can result in a shift in the distribution of states seen when performing pathfinding with the trained network versus the target network. Therefore, there is also the option to use the network currently being trained to guide pathfinding. However, this requires synchronizing between the parallel processes and the main processes, which slows both down. Furthermore, this requires double the DNN applications since, when guiding pathfinding with a target network, the nodes and edges required for the reinforcement learning update have already been evaluated with the target network during pathfinding. An overview of the parallelization of training is shown in Figure \ref{fig:trainmulti}. There is a special case of training where, instead of using reinforcement learning to compute targets, supervised learning is performed on the path costs of random walks. This has been shown to be effective at learning heuristic functions for puzzles such as the Rubik's cube \cite{chervovmachine}.

\subsection{Reinforcement Learning}
Problem instances are generated with a given parameter, $K$, where each problem instance is generated with a value, $k$, where $k$ is uniformly distributed between 0 and $K$ and $k$ denotes the length of the random walk used to generate the problem. However, user-defined implementations of sampling problem instances can ignore $k$ if deemed necessary. Every $U$ iterations, there is a check to see if the parameters of the target network, $\theta^-$, should be updated to the current parameters, $\theta$. This check can always return true or be based on a threshold for the training loss. During the first $U$ training iterations, the target network is a function that returns zero for all inputs.

Given a training batch size, $N$, $U\cdot N$ training data instances are generated per update check. Search with a given pathfinding algorithm is then performed for a maximum of $I$ iterations\footnote{Currently, the batch or beam size of the pathfinding algorithm used must be $1$ during training.}. All nodes expanded (for heuristic-v functions) or edges traversed (for heuristic-q functions) during search are used for training. Therefore, $\frac{U\cdot N}{I}$ problem instances are generated and, if a problem instance, $i$, generated with random walk length, $k_i$, is solved, then a new problem instance is generated in its place using the same $k_i$. Since a good value for $K$ may not be known beforehand, \PkgName{} has the option to automatically adapt $K$ during training by starting with $K=1$ and doubling it when 50\% of instances are solved. This is done until $K$ reaches some given $K_{\text{max}}$. 

After search, the corresponding reinforcement learning update is performed on the expanded nodes (value iteration), if training a heuristic-v function or traversed edges (q-learning), if training a heuristic-q function. The user may choose to use LHBL \cite{hadar2026beyond}, which recursively backs up the entire search tree before computing the target. LHBL has been shown to help learn heuristic functions that solve problems with fewer search iterations and better avoid depression regions \cite{aine2016multi}. \PkgName{} has the ability to learn policy functions by matching the distribution of edges traversed during pathfinding. However, this is still in the preliminary stages and may change significantly in the future. 

\begin{figure}[h]
	\centering
	\includegraphics[width=0.45\textwidth]{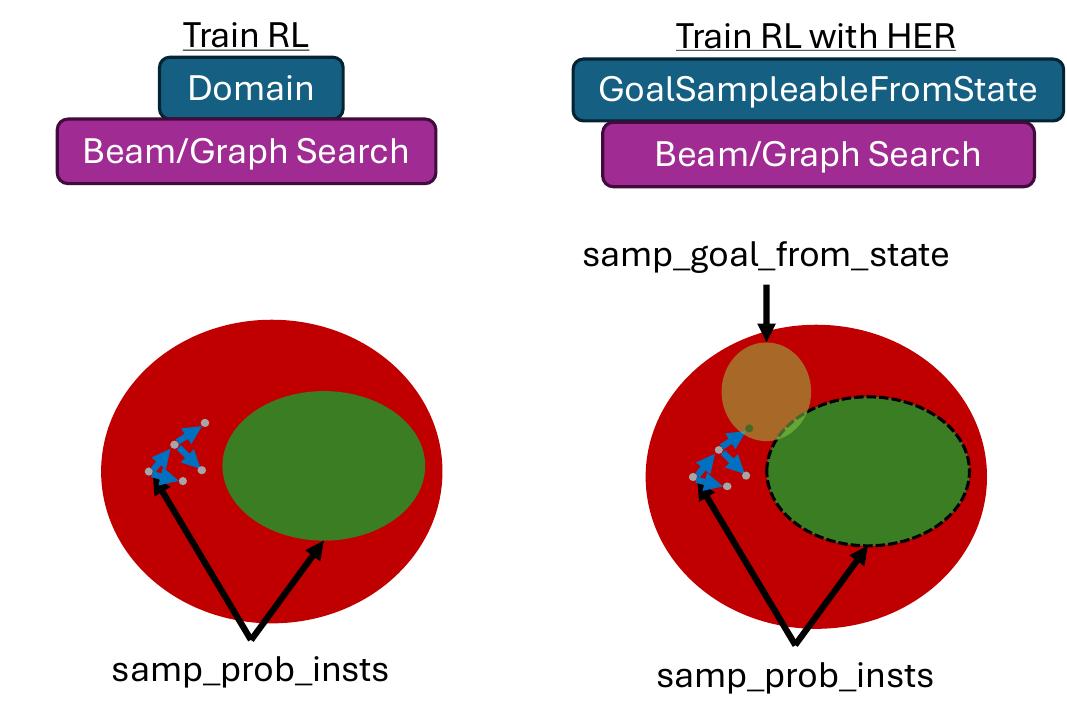}
	\caption{The two RL training approaches. For both approaches, problem instances are generated and the selected pathfinding algorithm is used to try to reach the goal from the start state. When using HER, if a path to the goal is not found, a goal is sampled from the state associated with a deepest node in the search tree and is used in place of the original goal.}
	\label{fig:trainrl}
\end{figure}

The user may choose to use hindsight experience replay (HER) \cite{andrychowicz2017hindsight} where, if the pathfinding algorithm fails to solve the problem in $I$ iterations, HER selects a deepest node in the search tree to generate a goal. This ensures that every search produces an example of a reached goal, which can facilitate learning when the generated problem instances are all difficult to solve. This training method also requires the domain to subclass \verb|GoalSampleableFromState|. A visualization of training without and with HER is shown in Figure \ref{fig:trainrl}. Lastly, a replay buffer may be used during training \cite{mnih2015human}, where the replay buffer stores the most recent nodes or edges along with their training targets from the past $R$ update checks and samples from the replay buffer to generate training data.

\subsection{Supervised Learning from Random Walks}
\begin{figure}[h]
	\centering
	\includegraphics[width=0.45\textwidth]{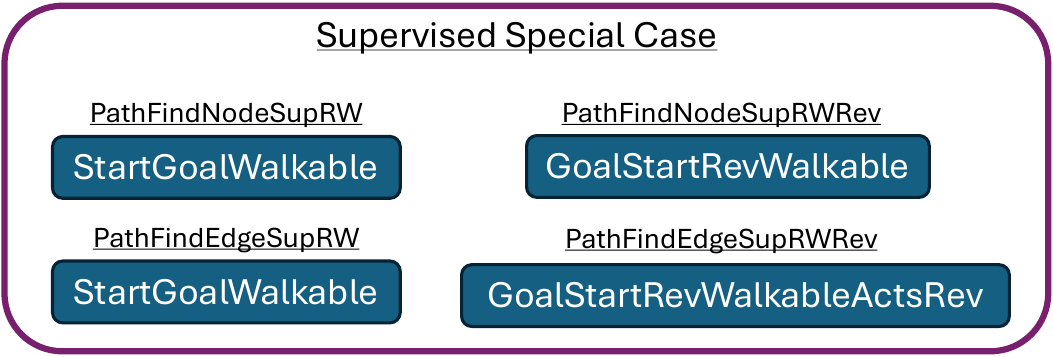}
	\caption{Pathfinding algorithms for the special case of supervised learning from path costs of random walks.}
	\label{fig:pathfindsup}
\end{figure}

Supervised learning from the path costs of random walks has also been shown capable of learning heuristic functions that perform well with large beam searches \cite{chervovmachine}. \PkgName{} can learn heuristic-v or heuristic-q functions with either random forward or reverse walks where the target is set to be the path cost of the random walk. One benefit to this approach is that no DNN is needed to compute targets and, therefore, targets can be obtained much faster. For heuristic-q functions learned with reverse walks, the domain must have reversible actions so, at the end of the reverse walk, the final edge on the reverse random walk that produces the start state can be reversed and used to update the heuristic-q function. \PkgName{} has the ability to learn policy functions by matching the distribution of edges traversed during the random walk. To function with the overall \PkgName{} architecture, these supervised learning approaches are implemented as pathfinding algorithms. A summary of these algorithms and their required mixin is shown in Figure \ref{fig:pathfindsup}.

\subsection{Monitoring Progress}
To monitor progress, TensorBoard can be used to plot the mean, minimum, and maximum values of the training targets, the training loss, the path cost and search iterations for solved problem instances and percentage of problem instances solved when generating training data using pathfinding as a function of training iteration. This pathfinding information is also plotted for test data, if a test dataset is given. \PkgName{} runs the given test pathfinding algorithms with the trained DNNs on the test dataset every given number of update checks.

\begin{figure}[h]
	\centering
	\includegraphics[width=0.45\textwidth]{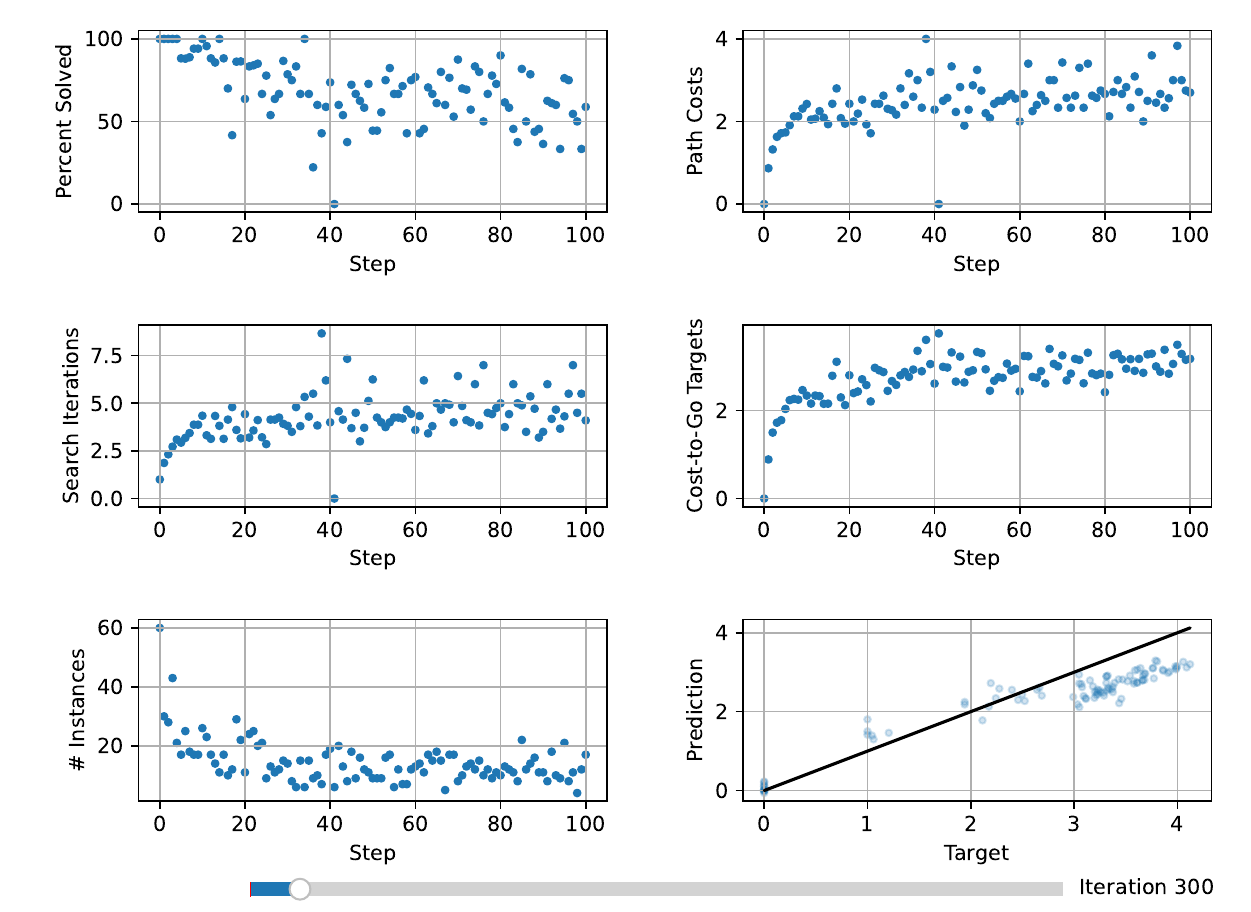}
	\caption{Visualization to monitor training progress.}
	\label{fig:trainsumm}
\end{figure}

Furthermore, a custom visualization can be accessed from the command line during and after training. This visualization shows the percentage of states solved, path costs, search iterations, the target cost-to-go, and the number of instances generated as a function of the random walk length used to generate problem instances. The visualization also shows a plot of the DNN predictions as a function of the target values. A slider can be used to see how these plots change as a function of training iteration, as shown in Figure \ref{fig:trainsumm}.

An output file is updated each update check that summarizes data generation, pathfinding performance, training performance, testing performance, and the time elapsed during execution of components. A verbose option is available that shows more detailed timing information for problem instance generation and pathfinding.

\section{Solving Pathfinding Problems}
Given a trained heuristic and/or policy function, any of the aforementioned pathfinding algorithms in Figure \ref{fig:pathfind} can be employed to solve problem instances. There is a verbose option for all pathfinding algorithms that prints the statistics of the path costs and heuristic values seen each iteration along with the frontier size (i.e. number of nodes in the beam or priority queue). \PkgName{} logs solutions found as well as important statistics such as the number of nodes generated, search iterations, and total search time in a Python pickle file. If the domain is a subclass of the \verb|StateGoalVizable| mixin, then the pickle file can be used to visualize solutions found during pathfinding.

If the pathfinding algorithm expects a heuristic function and is not given one, then it creates a heuristic function that always returns zero. This can be a convenient way to obtain a baseline using uniform cost search. If a policy function is expected and not given, then it creates a policy function that randomly samples actions with the \verb|samp_state_act| method.

\section{Other Mixins}
\subsection{NumPy Mixins}
For domains for which the next state method can be implemented in NumPy, \PkgName{} provides a \verb|NextStateNP| mixin where the user defines how to convert states to NumPy, compute the next states in NumPy, and convert NumPy to states. Based on these methods, the mixin implements the next state method and overrides the random walk method so that this conversion problem only happens once instead of every time the next state function is called. If the action space is enumerable, then the \verb|NextStateNPActsEnum| mixin overrides the expand function so that the conversion process only happens once.

\subsection{Answer Set Programming Mixin}
When representing states as a complete assignment of values to a set of variables, a goal can then be represented as a partial assignment of values to variables. Building on this, if an assignment is represented as a ground atom in first-order logic, then a goal can be represented with further abstraction by deriving logical predicates from these atoms. \PkgName{} provides the \verb|GoalGrndAtoms| mixin where states are represented as complete assignments and goals are represented as partial assignments. Derived predicates can then be defined with this mixin to be used to create answer set programs \cite{brewka2011answer} that represent more abstract goals. Answer set solvers \cite{gebser2014clingo} can then be used to find partial assignments that represent a set of goal states. This has been shown to allow for expressive goal specification \cite{agostinelli2024specifying} and allow for efficient goal reaching using answer set solvers and exploiting negation as failure through methods such as conflict-driven goal reaching \cite{agostinelli2025conflict}.

\subsection{PDDL Mixin}
One may wish to compare solvers from \PkgName{} with other solvers based on the planning domain definition language (PDDL) \cite{mcdermott20001998}, such as Fast Downward \cite{helmert2006fast}. The \verb|SupportsPDDL| mixin allows the user to define how to convert the \PkgName{} domain to a PDDL domain, a problem instance to a PDDL problem instance, as well as convert a string representation of a PDDL action to the \PkgName{} Python action representation.

\section{Convenience Features}

\subsection{Timing}
When implementing a domain, one may wish to time the methods in the domain as well as set breakpoints for debugging. \PkgName{} provides a tool that, given a domain, runs basic functionality such as generating problem instances, sampling actions, getting next states, and goal tests. It also runs more specific functionality, such as expand, based on what mixins the domain subclasses. If given a particular heuristic or policy DNN, it generates data to give to the DNN and times the execution. The user can set breakpoints anywhere in the domain or DNN code to help with debugging.

\subsection{Problem Instance Generation}
Given a domain, \PkgName{} can generate a given number of problem instances using a given minimum and maximum random walk length and save them to a file. If the domain allows for visualization, one can visualize the generated problem instances. One can also run uniform cost search, which requires no training, on the generated problem instances.

\section{Command-Line Tool}
Installing \PkgName{} also installs the \verb|deepxube| command-line tool. Given an implementation of a domain and DNN, all the aforementioned functionality can be accessed using the command line. \PkgName{} uses a registry system of domains, neural network inputs, and DNNs. Neural network inputs are associated with a domain when registered whereas DNNs are not, allowing a single DNN architecture to be used across multiple domains. Using the registered name, domains and DNNs can be referenced using the command line. Furthermore, a parser can be registered for domains and DNNs to allow one to specify domain arguments or DNN hyperparameters via the command line. This can make for convenient hyperparameter search. Multiple neural network inputs can be defined and registered for a single domain. When given a domain and DNN architecture, \PkgName{} searches for its corresponding neural network input for that domain using the registry and uses the first one it finds. Pathfinding algorithms and their arguments can also be given via the command line. For example, \verb|graph_q.10B_0.5W| performs BWQS with a batch size of 10 and weight of 0.5.

The \verb|deepxube| command-line tool expects a positional argument that determines the kind of functionality it will execute. For example, \verb|deepxube viz --domain cube3 --steps 5| visualizes a Rubik's cube that has been scrambled 5 times. All positional arguments can be accessed via \verb|deepxube --help| and all arguments for a particular positional argument can be accessed with \verb|--help| (e.g. \verb|deepxube viz --help|). The \verb|domain_info| and \verb|heuristic_info| show the information for registered domains and heuristic functions, respectively. \verb|time| times basic functionality and \verb|problem_inst| generates problem instances. \verb|train| trains DNNs, \verb|train_summary| shows the custom training progress visualization in Figure \ref{fig:trainsumm}, and \verb|solve| solves problem instances.

\section{Future Work}

\subsection{Policy Networks}
\PkgName{} currently supports the use of policy functions from which actions can be sampled in its pathfinding algorithms and has limited support for the training of such functions. Future work will build on algorithms in reinforcement learning for learning policies that can sample from arbitrary distributions \cite{haarnoja2017reinforcement,tessler2019distributional}. The ability to do so will make \PkgName{} more applicable to domains with continuous action spaces, such as those seen in robotics, as well as those with very large action spaces, such as when modeling multi-agent path finding as a joint action space \cite{tang2025railgun}.

\subsection{Learned Models}
Hand-coding a transition function for pathfinding problems can be impractical for real-world domains, such as robotic manipulation. Model-based reinforcement learning \cite{sutton1991dyna} with deep learning has been shown to be capable of learning transition functions based on observed transitions \cite{oh2015action,tianmodel}. Furthermore, the integration of learned models with planning has been shown to result in better performance than just using reinforcement learning alone \cite{asai2022classical,agostinelli2024learning}. Future work will allow for training of a transition function from observed transitions and using the trained transition function in the definition of the domain.

\subsection{Artificial Curiosity}
Besides user-defined methods to sample problem instances, \PkgName{} offers two mixins that both sample problem instances from random walks. While this has been observed to work well in practice \cite{agostinelli2019solving,tianmodel,chervovmachine}, there are many domains for which this will not produce useful problem instances. For example, for domains with many dead-ends, most random walks may end up in dead-ends and, therefore, not adequately explore the state space. Methods based on artificial curiosity or intrinsic motivation \cite{barto2004intrinsically} explore the state space based on some metric of novelty, such as next state prediction or inverse dynamics error \cite{pathak2017curiosity}. Related methods, such as learning to create a curriculum could be used to better explore the state space \cite{narvekar2020curriculum}.

\section{Conclusion}
The integration of deep learning, reinforcement learning, and heuristic search has proven capable of solving challenging problems, such as combinatorial puzzles and quantum algorithm compilation, and offers the potential of learning powerful domain-specific heuristic functions in a largely domain-independent fashion. \PkgName{} provides a package that applies state-of-the-art heuristic learning and heuristic search algorithms based on a user-provided definition of a pathfinding domain and DNN architecture. \PkgName{} provides convenient visualization and progress monitoring tools for implementing domains and DNNs, training DNNs, solving problems, and visualizing solutions. Pathfinding algorithms integrate seamlessly with DNNs and can take advantage of GPU parallelism with a large beam or batch size. Time-consuming aspects of training are parallelized across both CPUs and GPUs. \PkgName{} also provides functionality to speed up domain methods with NumPy as well as for expressive goal specification with answer set programming. Finally, this functionality is conveniently accessible via the \verb|deepxube| command-line tool.

\section*{Acknowledgements}
This material is based upon work supported by the National Science Foundation under Award No. 2426622.

\bibliographystyle{plainnat} 
\bibliography{mybib}

\end{document}